\documentclass[a4paper]{llncs}
 
 \newcommand\citep\cite 
 \newcommand\citet\cite

\newcommand\lpl{\mathtt{lpl}}

\newcommand\lang{\mathcal{L}}

\renewcommand\phi\varphi

\newcommand\LLL{\ensuremath{\mathcal{L}}}

\usepackage{graphicx}
\DeclareGraphicsExtensions{.pdf}

\newcommand\Land{\&^\Pi}

\newcommand\flex{ \textsc{(f)}}  
\newcommand\rig{ \textsc{(r)}} 

\newcommand\type[1]{^{#1}}
\newcommand\et\& 
\newcommand\fl{\rightarrow}

\usepackage{latexsym} 
\usepackage{amssymb} 

\newcommand{\ltyn}{\ensuremath{\Lambda\mathsf{Ty}_n}}
\newcommand\ma[1]{``\emph{#1}"}

\newcommand\ttt{\mathbf{t}}
\newcommand\eee{\mathbf{e}}

\newcommand\systF{\mbox{\textsf{F} }}

\usepackage{gb4e,cgloss4e} 

\newcommand\us{}

\author{Christian Retor\'e} 
\institute{LaBRI, Universit\'e de Bordeaux\\ (\& MELODI, IRIT-CNRS, Toulouse)} 
\title{Typed Hilbert Epsilon Operators\\ and the Semantics of Determiner Phrases\\ -- invited lecture --} 

\begin{document}

\maketitle 

\begin{abstract} The semantics of determiner  phrases, be they definite descriptions, indefinite descriptions or quantified noun phrases,   
is often assumed to be a fully solved question: common nouns are properties, and determiners are generalised quantifiers that apply to two predicates: the property corresponding to the common noun and 
the one corresponding to the verb phrase. 

We first present a criticism of this standard view.  Firstly, the semantics of determiners does not follow  the syntactical structure of the sentence. 
Secondly the standard interpretation of the  indefinite article
cannot account for nominal sentences. Thirdly, 
the standard view misses the linguistic asymmetry between the two properties of a generalised quantifier.

In the sequel, we propose a treatment of determiners and quantifiers as Hilbert terms in a richly typed system that we initially developed for lexical semantics, using a many sorted logic for semantical representations. We present this semantical framework called the Montagovian generative lexicon and show how these terms better match the syntactical structure and avoid the aforementioned problems of the standard approach. 

Hilbert terms rather differ from choice functions in that there is one polymorphic operator and not one operator per formula. 
They also open an intriguing connection between the logic for meaning assembly, the typed lambda calculus handling compositionality and the many-sorted logic for semantical representations. Furthermore epsilon terms naturally introduce type-judgements and confirm the claim that type judgment are a form of presupposition. 
\end{abstract}

\section{Presentation}

Determiners and quantifiers are an important ingredient of (computational) semantics, at least of the part of semantics known as formal semantics or compositional semantics,  that is concerned with what is asserted,  especially by a sentence:  such a semantical analysis tells \ma{who does what}
in a sentence. 

Researchers in formal linguistics, must be aware that semantics also includes other aspects like 
lexical semantics, distributional semantics, vectors of words for which there exist  
far more efficient natural language processing tools. 
These aspects of semantics rather concern \emph{what a text speaks about}. 

Of course both aspect are needed to understand the meaning, both for our human use of language and for the design of applications in natural language processing, like question answering by web searching. For instance, if one wants to know which guitar(s) played a rock star during a concert, the negation makes it difficult to extract the wanted information: 

\begin{exe}
\ex
\begin{xlist} 
\ex 
\emph{Question: Which guitars did he play at the concert.}  
\ex 
Funny he didn't play a Fender at that concert at least for one song. (web) 
\end{xlist}
\end{exe}

The standard treatment of determiners and quantifiers is to view them as generalised quantifiers, i.e. as functions of two predicates. 
In this paper we  argue that although such an account \ma{works} it is not really satisfactory mainly because it does not provide determiners with a proper logical form that can be interpreted on its own (as in the nominal phrase \ref{cars}, or when we just hear the indefinite noun phrase of example \ref{philosophy}) that would follow syntax (in example \ref{sang} generalised quantifiers require a predicate \ma{Keith sang \_} which does not correspond to any constituent)---  furthermore in the case of indefinite determiners it introduces a misleading symmetry between topic (theme) and comment (rheme) as example \ref{smokers} shows: these sentences do not speak about the same group.

\begin{exe}
\ex \label{cars} 
Cars, cars, cars....\footnote{Unless otherwise stated examples are from the Web} 
\ex \label{philosophy} 
\begin{xlist} 
\ex 
Some philosophy students ....
\ex 
\textit{We already have some image(s) in mind.}  
\ex 
Some philosophy students are "free spirits" who travel, read, and seek to live a non-traditional life.
\end{xlist} 
\ex \label{sang} 
Keith sang a song I never heard of. 
\ex \label{smokers} 
\begin{xlist}
\ex 
Some professors are smokers. 
\ex 
Some smokers are professors.
\end{xlist} 
\end{exe}

\section{The standard logical form of determiners}

The idea of Montague semantics is to map sentences to formulae of higher order logic (their logical forms)
in a way which implements the Fregean principle of compositionality:
typed functions (lambda terms) associated with words in the lexicon are composed according to the syntax.  
The glue logic is simply typed lambda calculus, over two types, $\eee$ for entities or individuals and $\ttt$ for propositions (that may there after be endowed with a truth value).

These typed lambda terms use two kinds of constants: connectives and quantifiers on the one hand and  individual constants and $n$-ary predicates  for the precise language to be described --- 
for instance a binary predicate like $delighted$ has the type $\eee\fl\eee\fl\ttt$.

\begin{figure}[t]
\begin{center} 
$
\begin{array}[t]{r|l}
\mbox{Constant} & \mbox{Type}\\ \hline 
	\exists & (\eee \fl \ttt) \fl \ttt \\ 
	\forall & (\eee \fl \ttt) \fl \ttt \\ 
\end{array} 
$ \hfill 
$
\begin{array}[t]{r|l} 
\mbox{Constant} & \mbox{Type}\\ \hline 
	\textrm{not} & \ttt \fl \ttt \\ 
	\textrm{and} & \ttt \fl (\ttt \fl \ttt) \\ 
	\textrm{or} & \ttt \fl (\ttt \fl \ttt) \\ 
	\textrm{implies} & \ttt \fl (\ttt \fl \ttt)
\end{array} 
$ \hfill 
$
\begin{array}[t]{r|l} 
\mbox{Constant} & \mbox{Type}\\ \hline 
	\mathit{played, sang} & \eee \fl (\eee \fl \ttt) \\ 
	\mathit{song} & (\eee \fl \ttt) \\ 
	\mathit{Keith} & \eee 
\end{array} 
$
\end{center}
\caption{Logical constants and language constants} 
\end{figure}

A small example goes as follows. Assume the syntax says that the structure of the sentence "\emph{Keith sang a song.}" is 
\begin{center}
(a (song))($\lambda y$ Keith sang $y$)
\end{center} 
where the function is always the term on the left. 
On the semantical side, this means that \ma{sang} is applied first to the property of \ma{being a song}  and to the property \ma{was sung by Keith}. 
If the semantical terms are as in the lexicon in Figure \ref{semanticlexicon}, placing the semantical terms in place of the words yields a large $\lambda$-term that can be reduced: 

\begin{figure}[b] 
\caption{A simple semantical lexicon} 
\label{semanticlexicon}
\begin{center} 
\begin{tabular}{ll} \hline 
\textbf{word} &  \textbf{\itshape semantical type $u^*$}\\ 
& \textbf{\itshape  semantics~: $\lambda$-term of type $u^*$}\\ 
&  {\itshape  $x\type{v}$ the variable or constant $x$ 
is of type $v$}\\ \hline 
\textit{a} 
& $(\eee\fl \ttt)\fl ((\eee\fl \ttt) \fl \ttt)$\\ 
& $\lambda P\type{\eee\fl\ttt}\  \lambda Q\type{\eee\fl\ttt}\  
(\exists\type{(\eee\fl \ttt)\fl \ttt}\  (\lambda z\type{\eee}  (\et\type{\ttt\fl (\ttt\fl \ttt)} (P\ z) (Q\ z))))$ \\  \hline 
\textit{song}  & $\eee\fl \ttt$\\ 
& $\lambda x\type{\eee} (\texttt{song}\type{\eee\fl \ttt}\  x)$\\  \hline 
\textit{sang} & $\eee\fl (\eee \fl \ttt)$\\ 
& $\lambda y\type{\eee}\  \lambda x\type{\eee}\  ((\texttt{sang}\type{\eee \fl (\eee \fl \ttt)}\  x)\  y)$ \\  \hline 
\textit{Keith} &$\eee$ \\ &  \texttt{Keith} 
\end{tabular}
\end{center} 
\end{figure}

$$
\begin{array}{c} 
\Big(\big(\lambda P\type{\eee\fl \ttt}\ \lambda Q\type{\eee\fl \ttt}\  (\exists\type{(\eee\fl \ttt)\fl \ttt}\  (\lambda z\type{\eee}  (\et (P\ z) (Q\ z))))\\ 
(\lambda u^\eee. \texttt{song}(u))
\big)
(\lambda y\type{\eee} (\texttt{sang}\type{\eee\fl \ttt}\  \texttt{Keith}) y)\Big) \\ 
\multicolumn{1}{c}{\downarrow \beta}\\ 
\lambda P\type{\eee\fl \ttt}\ \lambda Q\type{\eee\fl \ttt}\  (\exists\type{(\eee\fl \ttt)\fl \ttt}\  (\lambda Z\type{\eee}  (\et ((\lambda u^\eee. \texttt{song}(u))\ z)\\ ((\lambda y\type{\eee} (\texttt{sang}\type{\eee\fl \ttt}\  \texttt{Keith})\ y)\ z))))
 \\ 
\multicolumn{1}{c}{\downarrow \beta}\\ 
\big(\exists\type{(e\fl t)\fl t}\  (\lambda y\type{e}  (\et (\texttt{song}\type{e\fl t}\  y) ((\texttt{sang}\type{e\fl (e\fl t)}\  \texttt{Keith})\  y)))\big)
\end{array}
$$ 

This $\lambda$-term of type $\ttt$ that can be called the \emph{logical form} of the sentence, represents the following formula of predicate 
calculus (admittedly more pleasant to read): 

$$\exists y.\ (\texttt{song}(y)\ \et\ \texttt{sang}(\texttt{Keith},y))$$

This algorithm actually works because of the following result: 

There is a one to one correspondence between:
\begin{itemize}
\item 
the first order formulae over a first (respectively higher order) order language 
\LLL 
\item 
the closed normal lambda terms of type $\ttt$ with constants that correspond to connectives, quantifiers and  to the constants, functions and predicates in  \LLL. 
\end{itemize} 

The computation of the semantics of a sentence  boils down to complete the following steps (see e.g. \cite[Chapter 3]{MootRetore2012lcg}): 
\begin{enumerate} 
\item Parse the sentence, and turn the syntactic structure into  a (linear)  lambda term of type $\ttt$ (at least a functor argument structure, that is a binary tree with words as leafs and internal nodes specifying which subtree applies to the other one). This step is much easier when syntax is handled with categorial grammars.  
\item Insert at each word's place the corresponding semantical lambda term provided by the lexicon. 
\item Beta reduce this lambda term, the normal form being a logical formula, the semantical representation of the sentence. 
\end{enumerate}

\subsection{Some syntactical inadequacies of  the standard semantics of determiners} 
\label{struct} 

As noted in the introduction, the standard approach to determiners that we just recalled, is not fully satisfactory, and there are at least three reasons to be disappointed by the standard semantical analysis.

A first point is that when one hears a determiner phrase, he does not need  a complete sentence nor the main clause predicate to interpret the determiner phrase. 
This is easily observed from introspection: the simple utterance of a determiner phrase already suggests some interpretations, and possible referents, 
and references as individuals (sets of individuals, generic individual). It can also be observed in corpora: novels do include sentences without verbs. This can be observed in examples \ref{cars}, \ref{philosophy} above or in the following examples: when one reads \ma{some students}, he has an idea, an image in mind, 
as well as when he reads \ma{What a thrill} or \ma{an onion}. 
\begin{exe} 
\ex Some students do not participate in group experiments or projects.
\ex What a thrill --- My thumb instead of an onion. (Sylvia Plath) 
\end{exe}

A second point is that this formalisation misses the  asymmetry between the noun and the main clause predicate in existential statements. 
This asymmetry is the asymmetry between theme (or topic) and rheme (or comment) vanishes because both are assumed to be predicates
and the indefinite determiner simply asserts that something has both properties, and this \ma{and} is commutative. Even when both statements are felicitous, their meanings do differ: the sentence and its mirror image do not speak about the same class of objects. In the first case \ref{employees} one sentence can be said when speaking about universities or education and the next one when speaking about a company. This difference is even more striking in the example \ref{crooks}: sentences like the first one can be read and heard (our example is from Internet) while the second one or similar sentences cannot be found on the Internet: the reason is probably that \ma{crooks} do not really constitute a class one wants to speak about. 

\begin{exe} 
\ex \label{employees} 
\begin{xlist} 
\ex Some students are employees. 
\ex Some employees are students. 
\ex \label{crooks} 
\begin{xlist} 
\ex Some politicians are crooks.
\ex Some crooks are politicians. (no such examples on Internet)
\end{xlist}
\end{xlist} 
\end{exe} 

A third drawback is that the semantical or logical structure of the sentence does not match the syntactical structure (basically the parse tree) of the sentence. 
In the example we gave, this is patent: no constituent, no phrase does correspond to $\lambda x. (sang (Keith)) x^\eee$. This is related to the fact that the determiner or quantifier does not apply to a single predicate to form some term that can be interpreted. 

\begin{exe}
\ex 
\begin{xlist}
\ex 
Keith played some Beatles songs. 
\ex 
syntax (Keith (played (some (Beatles\ songs))))
\ex 
semantics: (some (Beatles songs)) ($\lambda x^\eee.$ Keith played $x$)
\end{xlist} 
\end{exe}

\subsection{Quantification and lexical semantics require a many sorted logic} 
\label{sorted} 

Let us point out that this Fregean view with a single sort prevents a proper treatment of quantification. 
Frege managed to express universal quantifiers (determiners like \ma{each} or \ma{every}) 
and existential quantifiers like \ma{a} or \ma{some} restricted to a sort, set or type  $A$ by using the following equivalences:  

\begin{exe}
\ex 
\begin{xlist} 
\ex 
$\forall x\in M\ P(x) \quad \equiv \forall x\ (M(x)\Rightarrow  P(x))$ 
\ex 
$\exists x\in M\ P(x) \quad \equiv \exists x\ (M(x) \& P(x))$
\end{xlist}
\end{exe} 

This treatment does not apply to other quantifiers like percentage or vague quantifiers:

\begin{exe}
\ex 
\begin{xlist} 
\ex 
$\textrm{for a third of the } x\in M\ P(x) \quad \not\equiv \forall x\ (M(x)\Rightarrow  P(x))$ 
\ex 
$\textrm{for few}  x\in M\ P(x) \quad \lnot\equiv \exists x\ (M(x) \& P(x))$
\end{xlist}
\end{exe}

Furthermore, as said in the first point of the previous subsection,  we would like to have a logical form or a reference for determiner phrases, even though the main predicate is still to come. 

\begin{exe}
\ex 
\begin{xlist} 
\ex The Brits
\ex The Brits love Australia, more than any other country except their own, according to an online survey for London's Daily Telegraph.
\end{xlist} 
\ex 
\begin{xlist} 
\ex Most students. 
\ex Most students will still be paying back loans from their university days in their 40s and 50s. 
\end{xlist} 
\end{exe}

This question is related to lexical semantics: 
what classes are natural, 
what sorts do we quantify over, 
what can possibly be the comparison classes
 that have not been uttered, 
 what are the sorts of complement a verb admit,  what verbs can apply to  a given sort of objects or of subjects? 
Our treatment of determiner phrases takes place in a framework that we initially designed for lexical semantics. 
But let us first speak about an alternative view of determiners and quantifiers.

\section{Hilbert operators, quantifiers, and determiners} 
\label{Hilbert} 

After the quantifier \emph{the one and unique individual such that $P$ \ldots} introduced by Russell for definite descriptions,
Hilbert (with Ackerman and Bernays) intensively used \emph{generic} elements for quantification, the study of which culminated in the second volume of \emph{Grundlagen der Mathematik} \cite{HBvol2}.  It should be stressed that these operators are introduced and described here with natural language examples, which is not so common in Hilbert's writings. 
We shall first present the $\epsilon$ operator which recently lead to important work in linguistics in particular with von Heusinger's work. \cite{EgliHeusinger1995,Heusinger1997,Heusinger2004}

\subsection{An ancestor to Hilbert operators: Russell's iota for definite descriptions} 

The first step due to Russell was to denote by $\iota_x.\ F$ the unique individual enjoying the property $F$ 
in a definite description like the first sentence below and to remain undetermined when existence and uniqueness do not hold. 
\cite{Russell1905}

\begin{exe} 
\ex \emph{\emph{The present president of France} was born in Rouen.}\us \glt (existence and uniqueness hold) 
\ex \emph{\emph{The present king of France} was born in Pau.}\us  \glt (existence fails) 
\ex \emph{\emph{The present minister} was born in Barcelona.}\us  \glt (uniqueness fails) 
\end{exe} 

Of course this operator is not handy from a logical or formal  point of view since the negation of \ma{there exists a unique x such that $P(x)$}
  is \ma{either no $x$ or more than two $x$ enjoys $\lnot P$}: its negation is clearly inelegant and indeed there are no well behaved deduction rules for such an operator. 
However, as observed by von Heusinger the uniqueness even when using the \emph{definite} article is not really mandatory: it should refer to a salient element in the speaker's view, and in many examples the definite description is neither unique nor objectively salient, we shall come back to this point at the end of the present paper.

\subsection{Hilbert epsilon and tau} 

From this idea, Hilbert introduced  an individual existential term 
defined from a formula: given a formula $F(x)$ with a free variable $x$ 
one defines the term $\epsilon_x.\ F$  in which the occurrences of $x$ in $F$ are bound 
(this is the original notation, nowadays this term is often written as $\epsilon x.\ F$). 
Whenever \textbf{some} element, say $a$,  enjoys $F$, then the  epsilon term $\epsilon_x.\ F$ enjoys $F$.

Dually, Hilbert introduced a universal generic element $\tau_x.\ F$, which corresponds to the generic elements used in mathematical proofs:
to establish that a property $P$ holds for every integer, the proof usually 
starts with \ma{Let $n$ be an integer, \ldots}  where $n$ has no other property than being an integer.  
Consquently when this generic integer has the property, so does any integer. 
The $\tau$-term  $\tau_x.\ F$ is the dual of the $\epsilon$-term $\epsilon_x.\ F$
: $\tau_x.\ F$ enjoys the property $F$ when \textbf{every} individual does.

More formally, given a first language $\lang$ (constants, variables, function symbols, relation symbols, the later two with an arity) here is a precise definition of the epsilon terms and formulae. 
Terms and formulae are defined by mutual recursion:

\begin{itemize} 
\item 
Any  constant in $\lang$ is a term.
\item 
Any  variable in $\lang$ is  a term.
\item 
$f(t_1,\ldots , t_p)$ is a term provided each $t_i$ is a term and $f$ is a function symbol of arity $p$ 
\item 
$\epsilon_x A$ is a term if $A$ is a formula and $x$ a variable and  any free occurrence of $x$ in $A$ is bound by $\epsilon_x$ 
\item 
$\tau_x A$ is a term if $A$ is a formula and $x$ a variable and  any free occurrence of $x$ in $A$ is bound by $\tau_x$ 
\item 
$s = t$ is a formula whenever $s$ and $t$ are terms. 
\item 
$R(t_1,\ldots,t_n)$  is a formula  provided each $t_i$ is a term and $R$ is a relation symbol of arity $n$ 
\item 
$A \& B$, $A \lor B$, $A \Rightarrow  B$ are formulae if $A$ and $B$ are formulae 
\item 
$\lnot A$ is formula if $A$ is a formula. 
\end{itemize}

As the example below shows, a  formula of first order logic can be recursively  translated into a formula of the epsilon calculus, without surprise. Admittedly the epsilon translation of a usual  formula may  look quite  complicated --- at least we are not used to them:\footnote{We shall not use such formulae as semantical representations: indeed, they are even further away from the syntactical structure than usual first order formulae.}

\newcommand\epsi{\epsilon_y P(\tauu,y)}
\newcommand\tauuu{\tau_x P(x,\epsi)}
\newcommand\tauu{\tau_x P(x,y)}

\begin{exe}
\ex 
\begin{xlist} 
\ex 
$\forall x\ \exists y\ P(x,y)$
\ex 
$= \exists y\ P(\tauu, y)$  
\ex 
$= P(\tauuu,\epsi)$
\end{xlist} 
\end{exe}

The deduction rules for $\tau$ and $\epsilon$ are the usual rules for quantification: 
\begin{itemize} 
\item 
From $A(x)$ with $x$ generic in the proof (no free occurrence of $x$ in any hypothesis),  infer $A(\tau_x.\ A(x))$ 
\item 
From $B(c)$ infer $B(\epsilon_x.\ B(x))$.  
\end{itemize} 

The other  rules can be found by duality: 
\begin{itemize} 
\item 
From $A(x)$ with $x$ generic in the proof (no free occurrence of $x$ in any hypothesis),  infer $A(\epsilon_x.\ \lnot A(x))$
\item 
From $B(c)$ infer $B(\tau_x.\ \lnot B(x))$ 
\end{itemize} 

Hence we have: 
\begin{eqnarray*}
F(\tau_x.\ F(x))&\equiv& \forall x. F(x)\\ 
F(\epsilon_x.\ F(x))&\equiv& \exists x.\ F(x)\\[1ex] 
\tau_x. A(x)  &=& \epsilon_x. \lnot A(x) 
\end{eqnarray*}

Because of the latest equation due to the classical negation ($\forall x.\ P(x)\equiv \lnot \exists x.\ \lnot P(x)$), only one of these two operators $\tau$ and $\epsilon$ is needed: commonly people choose  the $\epsilon$ operator.

This logic is known as the \emph{epsilon calculus}.

Hilbert turned these symbols into a mathematically satisfying theory, since it allows to fully 
describe quantification with simple rules. The first and second epsilon theorem 
basically say that this is an alternative 
formulation of first order logic. 
\begin{description}
\item[First epsilon theorem] 
When inferring a formula $C$ without the $\epsilon$ symbol nor quantifiers from  formulae $\Gamma$ not involving the $\epsilon$ symbol nor quantifiers  the derivation can be done within quantifier free  predicate calculus. 
\item[Second epsilon theorem]
When inferring a formula $C$ without the $\epsilon$ symbol from  formulae $\Gamma$ not involving the $\epsilon$ symbol,   the derivation can be done within usual predicate calculus. 
\end{description}

In this way, Hilbert provided the first correct proof of Herbrand's theorem (much before mistakes where found and solved by Goldfarb) 
and a way to prove the consistence of Peano's arithmetic at the same time as Gentzen did. 

Later on Asser \cite{Asser1957} and Leisenring \cite{Leisenring1967epsilon} have been working on epsilon calculus in particular for having models and completeness, and for cut-elimination.  Nevertheless, as one reads on \emph{Zentralblatt \textsc{math}} these results are misleading as well as the posterior corrections --- see in particular 
\cite{Zbl0327.02013,Zbl0381.03042} and the related reviews. 
Only the proof theoretical aspects of  the epsilon calculus seem to have been further investigated with some success  in particular by Moser and Zach \cite{MoserZach2006etheorem} and Mints  \cite{mints2008}. \footnote{While correcting these lines before printing, we just learnt that this great logician Grigori (Grisha) Mints passed away; sincere condolences to his family, friends and to the logic community.}

\subsection{Hilbert's operators in natural language} 

In Hilbert's book the operators $\epsilon$ and $\tau$ are explained with natural language examples,
but a very important and obvious linguistic property is not properly stated:  
the $\epsilon_x F$  has the type (both in the intuitive and in the formal sense) of a noun phrase, 
and is meant to be the argument of a predicate (for instance the subject of a verb),
thus being a \emph{suppositio} in the medieval sense. 
\cite{libera1996querelle,KK86}

Nowadays there has been a renewed interest in the epsilon formulation of quantification,
in particular by von Heusinger. He uses a variant of the epsilon for definite descriptions, leaving out the uniqueness of the iota operator of Russell, one reason being that the context often determines a unique object, the most salient one. 
We call it a ``variant" because it is not clear whether one still has the equivalence with ordinary existential quantification: von Heusinger constructs an epsilon term whenever there is an expression like \emph{a man} or \emph{the man}  but it is not clear how 
 one asserts that $man(\epsilon_x.\ man(x))$. The distinction between $\epsilon$ and $\eta$ is that the former selects the most salient possible referent, while the later selects a new one.

\subsection{Hilbert's operators, beyond usual logic}
\label{beyond} 

The study of epsilon operators focused on usual logic, typically first order classical logic within this extended language. 
Epsilon and the epsilon substitution method were part of Hilbert's program to provide finistic consistency proofs for arithmetic (and even analysis, using second order epsilon).  Hence, although by that time people were probably aware that it goes beyond usual first order, none spoke about this extension. 

Here is an extremely simple example of a formula of the epsilon calculus without an equivalent in first order logic, that von Heusinger and us use for natural language semantics as explained below: 
 
 $$F=P(\epsilon_x Q(x))$$ 
 
This formula, according to the aforementioned epsilon rules,  entails $G=P(\epsilon_x P(x))$ (i.e. $\exists x.\ P(x)$), 
but it does not entails $H=Q(\epsilon_x Q(x))$ (i.e. $\exists x.\ Q(x)$). 
Of course, if one further assumes $H$, then the  formulae $F$ and $H$ entail, according to epsilon rules, 
the $P\& Q(\epsilon_x.\ P\&Q)$ that is $\exists x.\ P\& Q(x)=\exists x. P(x)\& Q(x)$. But there is no first order formula equivalent to this simple epsilon formula $F$.

\section{Determiners in the Montagovian generative lexicon}

The standard view in Montague semantics is in perfect accordance with Frege's view of entities: a single universe gathers all entities. 
Hence a definite or indefinite determiner picks one element from this single sorted universe and a quantifier ranges over this single universe.
As said in subsections \ref{sorted} and \ref{struct}, this view of quantification does not really match our linguistic competence nor our cognitive abilities. 

This question is related to another part of semantics, namely lexical semantics. If one wants to integrate some lexical issues in a  compositional framework,
one needs sorts or many base types for entities, in order to specify what should be the nature of the arguments of a given word.  This question is related to the type of the semantical constants: what should be the domain of a predicate,  what are the relations between these logical constants? 
Observe, for instance that in Montague semantics a verb phrase and a common noun have the very same type $\eee\fl\ttt$, 
that events are standard entities, and that there is no way to have privileged relation between predicates and arguments:
for instance a \ma{book} can be \ma{enjoyed, disliked, read, written, printed, bound, burnt, lost,...} 

As the two questions are linked, we here present a compositional framework for semantics that accounts for both lexical issues and for the present question of determiners and quantifiers. 

\subsection{The Montagovian generative lexicon} 
As observed above,  it would be more accurate to have many individual base types rather than just $\eee$. Thus, the 
application of a predicate to an argument may only happen when it makes sense. 
Some sentences should be ruled out like \ma{The chair barks.} or \ma{Their five is running.}, 
and this is quite easy when there are several types for individuals:  
the lexicon can specify \ma{barks} and \ma{is running}  
only apply to individuals of type \ma{animal}. 
Nevertheless, such a type system needs to incorporate some flexibility. Indeed, in the context of a football match, the second sentence makes sense:  \ma{their five}  can be the player wearing the 5 shirt and who, being \ma{human}, is an \ma{animal} that can \ma{run}.

Our system is called the Montagovian Generative Lexicon or $\Lambda Ty_n$. Its lambda terms extend the simply typed ones of Montague semantics above. Indeed, we use second order lambda terms from Girard's   
system 
$\systF$ (1971) 
\citep{Girard2011blindspot}.

The types of \ltyn are defined as follows: 
 \begin{itemize} 
\item 
Constants types $\eee_i$ and $\ttt$, as well as type variables $\alpha,\beta,\ldots$ are types. 
\item 
$\Pi\alpha.\ T$ is a type whenever $T$ is a type and $\alpha$  a type variable . The type variable may or may not occur in the type $T$. 
\item 
$T_1\fl T_2$ is a type whenever $T_1$ and $T_2$ are types. 
\end{itemize}

The terms of $\ltyn$, are defined as follows: 
\begin{itemize} 
\item A variable  of type $T$ i.e. $x:T$ or  $x^{T}$  is a \emph{term}, and there are countably many variables of each type.
\item In each type, there can be a countable set of constants of this type, and a constant of type $T$ is a term of type $T$. Such constants are needed for logical operations and for the logical language (predicates, individuals, etc.). 
\item 
$(f\ t)$ is a term of type $U$ whenever $t$ is a term of type $T$ and  $f$ a term of type $T\fl U$. 
\item 
$\lambda x^{T}.\ \tau$ is a term of type $T\fl U$ whenever $x$ is variable of type $T$, 
and $t$ a term of type $U$.  
\item $t \{U\}$ is a term of type $T[U/\alpha]$
whenever $\tau$ is a term of type $\Pi \alpha.\ T$, and $U$ is a type. 
\item $\Lambda \alpha. t$ is a term of type $\Pi \alpha. T$
whenever $\alpha$ is a type variable, and  $t:T$ a term without any free occurrence of the type variable $\alpha$ in the type of a free variable of $t$.  
\end{itemize}

The later restriction is the usual one on the proof rule for quantification in propositional logic: one should not conclude that $F[p]$ holds for any  proposition $p$
when assuming $G[p]$ --- i.e. having a free hypothesis of type $G[p]$. 

The reduction of the terms in system F or its specialised version \ltyn\ is defined by the two following reduction schemes that resemble each other:  
\begin{itemize} 
\item $(\lambda x. \tau) u$ reduces to $\tau[u/x]$ (usual $\beta$ reduction). 
\item $(\Lambda \alpha. \tau) \{U\}$  reduces to $\tau[U/\alpha]$ (remember that $\alpha$ and $U$ are types). 
\end{itemize} 

As  \citet{Girard71,Girard2011blindspot} showed 
reduction is strongly normalising and confluent 
\textit{every term of every type admits a unique normal form which is reached no matter how one proceeds.} 
This has  a good consequence for us, see e.g. \citep[Chapter 3]{MootRetore2012lcg}: 

\begin{quotation}\it 
\noindent \textbf{\ltyn\ terms as formulae of a many-sorted logic} 
If the predicates, the constants and the logical connectives and quantifiers 
are the ones  from a many sorted logic of order $n$ (possibly $n=\omega$) then the closed normal terms of $\ltyn$ of type $\ttt$ unambiguously correspond to many sorted  formulae of order $n$. 
\end{quotation} 

\begin{figure} 
\label{polyandfig} 
\begin{center}
\includegraphics[scale=0.25]{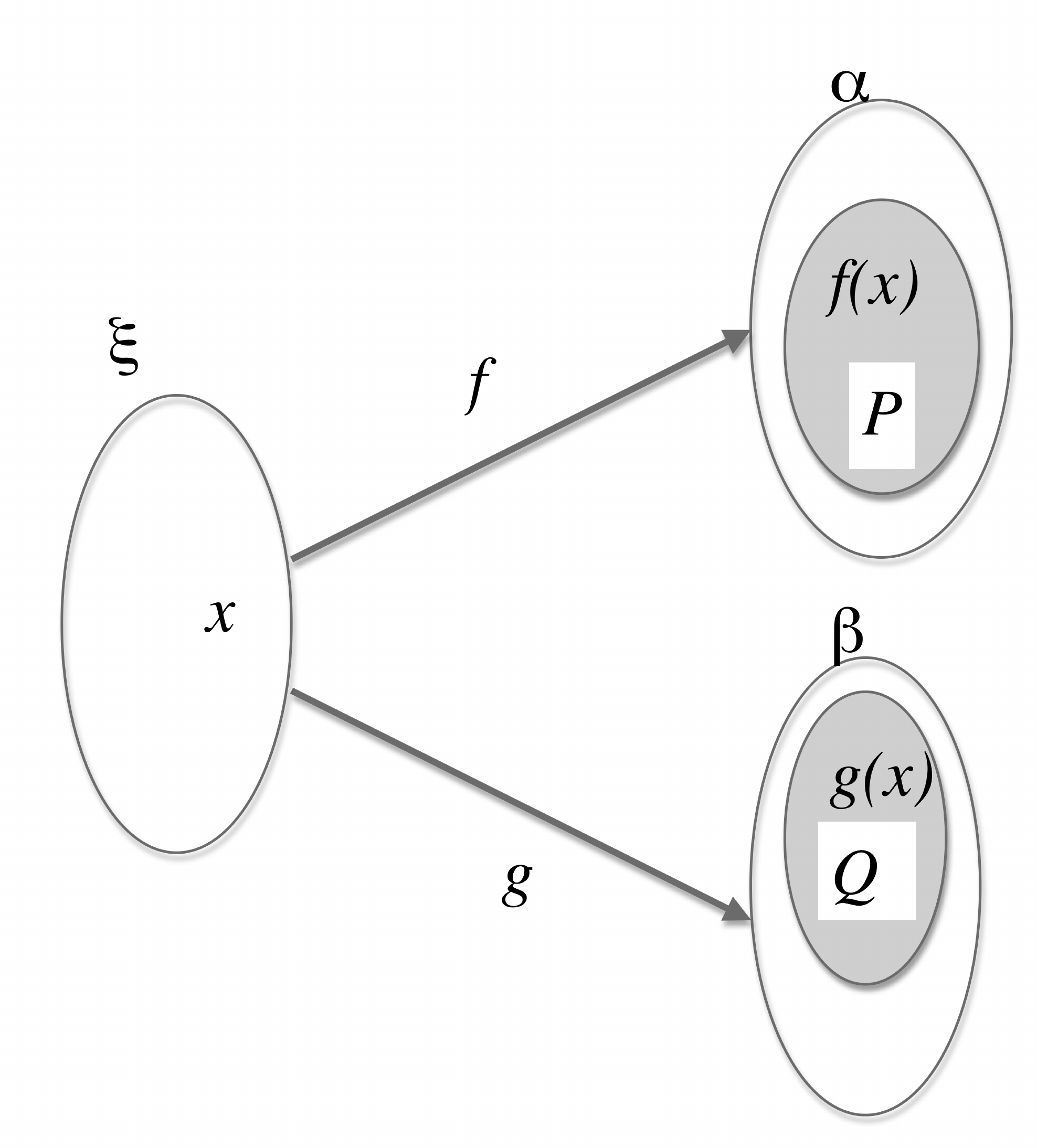} 
\end{center} 
\caption{\large Polymorphic and: $P(f(x))\& Q(g(x))$ [$x{:}\xi$, $f{:}\xi\fl\alpha$, $g{:}\xi\fl\beta$].}  
\end{figure}

Polymorphism allows a factored treatment of phenomena that treat uniformly 
families of types and terms. An interesting example is the polymorphic conjunction for copredication: \emph{whenever} 
an object $x$ of type $\xi$ can be viewed both:
\begin{itemize}
\item  as an object 
of type $\alpha$ (via a term $f_0:\xi\fl\alpha$) to which a property $P^{\alpha\fl\ttt}$ applies 
\item and as an object of type $\beta$  
to which a property $Q^{\beta\fl\ttt}$ applies
(via a term $g_0:\xi\fl\beta$), 
\end{itemize} 
the fact that 
$x$ enjoys  $P\& Q$ can be expressed by the unique polymorphic term (see explanation in figure \ref{polyandfig}): 
\begin{exe} 
\ex \label{polyandterm} 
$\Land=\Lambda \alpha \Lambda \beta
\lambda P^{\alpha \fl \ttt} \lambda Q^{\beta\fl \ttt} 
 \Lambda \xi \lambda x^\xi 
 \lambda f^{\xi\fl\alpha} \lambda g^{\xi\fl\beta}.\\ 
\hspace*{15em}\hfill (\et^{\ttt\fl\ttt\fl\ttt} \ (P \ (f \ x)) (Q \ (g \  x))) 
$
\end{exe} 

The lexicon provides each word with:

\begin{itemize} 
\item 
A main $\lambda$-term of \ltyn, the ``usual one" specifying the argument structure of the word.
 \item 
A finite number of $\lambda$-terms of \ltyn\ (possibly none) that implement meaning transfers. Each meaning transfer is declared in the lexicon to be \emph{flexible} \flex\ or \emph{rigid} \rig. 
\end{itemize}

\begin{figure} 
\caption{A sample lexicon}
$$ 
\begin{array}{l|l|rl} 
\mbox{word} & \mbox{principal\ $\lambda$-term} & \multicolumn{1}{l}{\mbox{optional\ $\lambda$-terms}} & \mbox{rigid/flexible}\\ \hline 
Liverpool & \lpl^T & Id_T:T\fl T &\flex \\ 
& & t_1:T\fl F &\rig \\ 
& & t_2:T\fl P &\flex \\ 
& & t_3:T\fl Pl &\flex\\ 
\hline 
spread\_out & spread\_out:Pl\fl\ttt & \\ 
\hline 
voted & voted:P\fl\ttt& \\
\hline 
won & won:F\fl\ttt&\\  
\end{array}
$$
where the base types are defined as follows: 
\begin{tabular}[t]{ll}
$T$ & town \\ 
$P$ & people \\ 
$Pl$ & place \\
\end{tabular} 
\label{lexicon} 
\end{figure}

Let us see how such a lexicon works. 
When a predication requires a type $\psi$ (e.g. Place) while its argument is of type $\sigma$ (e.g. Town)
the optional terms in the lexicon can be used to ``convert" a Town into a Place.

\begin{exe}
\ex \label{lplex}
\begin{xlist} 
\ex Liverpool is spread out. \label{ll} 
\ex 
This sentence leads to a type mismatch $spread\_out^{Pl\fl\ttt}(\lpl^T))$, since \ma{spread\_out} applies to \ma{places} (type $Pl$) and not to \ma{towns} as \ma{Liverpool}. 
This type conflict  is solved using the optional term $t_3^{T\fl Pl}$ provided by the entry for \ma{Liverpool}, which turns a town ($T$) into a place ($Pl$) 
 $spread\_out^{Pl\fl\ttt}(t_3^{T\fl Pl} \lpl^T))$ --- a single optional term is used, the \flex / \rig difference is useless. 
\end{xlist} 
\ex
\begin{xlist}
\ex 
Liverpool is spread\_out and voted (last Sunday).  \label{llv} 
\ex 
In this example, the fact that \ma{Liverpool} is \ma{spread\_out} is derived as previously, and the fact \ma{Liverpool} \ma{voted} is obtained from the transformation of the town into people, which can vote. The two can be conjoined by the polymorphic \ma{and} defined above  in \ref{polyandterm} ($\Land$) 
because these transformations are flexible: one can use both of them.  
We can make this precise using only the rules of our typed calculus. 
The syntax yields the predicate $(\Land (is\_spread\_out)^{Pl\fl \ttt} (voted)^{P\fl\ttt})$ and consequently 
the type variables should be instantiated by $\alpha:=Pl$ and $\beta:=P$ and the exact term is 

$\Land \{Pl\} \{P\} (is\_spread\_out)^{Pl\fl \ttt} (voted)^{P\fl\ttt}$ 

which reduces to:  

$ \Lambda \xi \lambda x^\xi  \ 
 \lambda f^{\xi\fl\alpha} \lambda g^{\xi\fl\beta}  
(\et^{\ttt\fl\ttt)\fl\ttt} \ (is\_spread\_out \ (f \ x)) (voted \ (g \  x)))$. 

Syntax also says this term is applied to \ma{Liverpool}. 
which forces the instantiation $\xi:=T$ and the term corresponding to the sentence is after some reduction steps,\\  
$ \lambda f^{T\fl Pl} \lambda g^{T\fl P}  
(\et \ (is\_spread\_out \ (f \ \lpl^T)) (voted \ (g \  \lpl^T))))$. Fortunately the optional $\lambda$-terms 
$t_2:T\fl P$ and  $t_3:T\fl Pl$ are provided by the lexicon, and they can both be used, since none of them is rigid.
Thus we obtain, as expected\\  
$(\et \ (is\_spread\_out^{Pl\fl\ttt} \ (t_3^{T\fl Pl} \ \lpl^T)) (voted^{Pl\fl\ttt} \ (t_2^{T\fl P} \  \lpl^T)))$ 
\end{xlist}
\ex 
\begin{xlist} 
\ex \label{lvw}    \#
Liverpool voted and won (last Sunday). 
\ex 
This third and last example is rejected as expected. Indeed, the transformation of the town into a football club prevents any other transformation (even the identity) to be used with 
the polymorphic \ma{and} ($\Land$) defined above  in \ref{polyandterm}. 
We obtain the same term as above, with $won$ instead of $is\_spread\_out$:\newline   $ \lambda f^{T\fl Pl} \lambda g^{T\fl P}  
(\et \ (won \ (f \ \lpl^T)) (voted \ (g \  \lpl^T))))$ 
and the lexicon provides the two morphisms that would solve the type conflict, but one of them is 
\emph{rigid}, i.e. we can solely use  this one. Consequently no semantics can be derived from this sentence,  which is semantically invalid. 
\end{xlist} 
\end{exe}

The difference between our system and those of  \citet{Luo2011lacl,Asher2011wow}  
does not come down to  the type systems, which are quite similar, 
but in the \emph{architecture} which  is, in our case,  rather \emph{word driven} than type driven. The optional morphisms are anchored in the words, and do not derive from the types. This is supported in our opinion by the fact that some words with the very same ontological type (like French nouns  \ma{classe} and \ma{promotion}, that are groups of students in the context of teaching) may undergo different coercions (only the first one can mean a classroom). 
This rather lexicalist view goes well with the present work that proposes to have specific entries for deverbals, 
that are derived from the verb entry but not automatically. 

This system has been implemented as an extension to the Grail parser \citet{moot10grail}, with $\lambda$-DRT instead of formulae as $\lambda$-terms. 
It works fine once the semantical lexicon has been typeset.\footnote{Syntactical  categories are learnt from annotated corpora, but  semantical typed $\lambda$-terms cannot yet be learnt, as discussed in the conclusion.}

We already explored some of  the compositional properties (quantifiers, plurals and generic elements,....) of our Montagovian generative lexicon as well as some of 
the lexical issues (meaning transfers, copredication, fictive motion, deverbals, ...  )  
\citep{BMRjolli,Retore2012rlv,Retore2013taln,MMR2013lenls,RealRetore2014jolli}. 

\subsection{Determiners as typed epsilon operators} 

As we saw there are many base  types that are sorts of the many sorted logic and even more complex types over which one may quantify, 
a fairly natural semantics for determiners is to pick one element in its sort. 

For instance, consider the  indefinite determiner \ma{a}. It should be seen as an operator 
acting on a noun phrase without determiners that outputs some individual. 
In order to make things correct and precise, consider the noun phrase, \ma{a cat} where \ma{a} 
acts upon \ma{cats},  and think about the possible types of \ma{a}, which 
clearly it depend on what \ma{cat} is. Is $cat$  a type or a property satisfied by \ma{cats} among a larger class or type? 
\begin{enumerate} 
\item 
If \ma{cat} is a type the constant  for \ma{a}  should be of type $\Pi\alpha.\ \alpha$. 
\item \label{iotaprop}
If \ma{cat} is a property, say of a larger type \ma{animal}, then this constant should take a property of animals of type $animal\fl\ttt$ and yield a cat. 
Now assume that the property is a more complex property $P$ \ma{cat which lives nearby}, what should \ma{a} do? It should apply to a property of animals like $P$  and yields an entity $x$ that enjoys $P$. Because $x$ enjoys $P$ its type should be \ma{animal}. 
In this case the type of the constant corresponding to $P$ should be $\Pi\alpha.\ (\alpha\fl\ttt)\fl \alpha$, hence the type does not guarantee by itself that 
$x$ enjoys $P$ and consequently a presupposition $P(a\ cat)$ has to be added. 
\end{enumerate} 

We deliberately chose to use option \ref{iotaprop} and only this one. Firstly,  we cannot avoid this case, because not every property that a determiner may apply to can be assumed to be a type, there would be too many of them. Secondly, the first option can be encoded within the second option. Indeed if there is a type $cat$ one can consider a predicate \ma{being a cat}. Indeed, unsurprisingly, the semantics of predicates and the one of quantifiers and determiners are closely related. 

Usually, a determiner or a quantifier applies to one (\ma{everyone}) or two (\ma{a}) predicates and yields a proposition. 
A Hilbert operator combines with \emph{one} predicate and yields a term, an entity. 
In a many sorted and typed system like $\ltyn$ what is the type of a predicate? 
The standard type for a predicate is $\eee\fl\ttt$, 
but given the many sorts $\eee_i$ we could have predicates that apply to other entity type than $\eee$. 
Is  \ma{cat} a property of individuals of type  \ma{animal} if such a type exists or is it a property that may apply to any entity, and which is constantly false outside of the type \ma{animal}? 
If the domain of a predicate is 
 $\eee_i$ and not $\eee$ (the type of all entities),   a predicate $P^{\eee_i\fl\ttt}$ canonically extends to a predicate $\overline{P}^{\eee\fl\ttt}$ by saying it never holds outside of $\eee_i$. Conversely a property like  $cat$ whose domain is some $\eee_i$ (e.g  \ma{animal}) 
 can be restricted to any subtype of $\eee_i$, but in case the subtype of $\eee_i$ does not include all \ma{cats} there dis no way to recover the initial predicate \ma{cat} that applies to animals. 
 
Now that we have a proper representation of a predicate in the type system, one may wonder how a type can be reflected as a predicate. 
For instance what should be the type of a predicate associated with a type, like \ma{being a cat} if \ma{cat} is a type. 
Natural domains for the such a predicate could be \ma{animals}, \ma{mammals}, \ma{felines},\ldots As it is difficult to chose,
let us decide that the domain of a given predicate associated with a type always is the largest, the collection of all possible entities $\eee$ which can be restricted as indicated above. Hence \ma{being of type $\alpha$} that we write 
$\widehat \alpha$ is of  type $\eee\fl\ttt$

So far we have not said what are the base type which intervenes in representing predicates and quantifiers.  We need several of them, to express selectional restrictions . Asher \cite{Asher2011wow} uses a dozen of ontological types (events, physical objects, human beings, information, etc.) 
Luo \citet{Luo2012lacl} suggests using a flat ontology with common nouns (there are thousands of them) as base types. With  Mery we suggested to consider classifiers (100--200) as in languages that have classifiers (sign language, Chinese, Japanese) \cite{MeryRetore2013nlpcs}.

As said above the lexicon associate the constant  $\epsilon$ of type  $\Pi\alpha.\ (\alpha\fl\ttt)\fl\alpha$ to the indefinite article 
--- that is an Hilbert/von Heusinger  $\epsilon$ adapted to the typed case. 
Hence the indefinite article is a polymorphic 
 $\epsilon$ that specialises to a type/sort  $\{\eee_i\}$ and applies to a predicate $P$ of type $\eee_i\fl\ttt$ yielding an entity of type $\eee_i$. 
Let us consider an extremely simple example:  
($ani$ stands for the type of animals): 

\begin{exe}
\ex \begin{xlist}
\ex A cat sleeps (under your car). 
\ex \label{chattermun} term for \ma{a}: $\epsilon:\Pi\alpha.\ ((\alpha\fl\ttt)\fl\alpha)$
\ex term for \ma{sleep}: $(\lambda x.\ sleeps^{ani\fl \ttt}(x))$
\ex term for \ma{cat}: $(\lambda x.\ cat^{ani\fl \ttt}(x))$
\ex \label{chatsynt} syntax: $((a \fl cat) \leftarrow sleeps)$ 
\ex \label{chatsem} semantics: $sleeps (a\ cat)$ 
\ex \label{chattermsem} $(\lambda x.\ sleeps^{ani\fl \ttt}(x)) (\epsilon^{\Pi\alpha.\ ((\alpha\fl\ttt)\fl\alpha)} cat^{ani\fl \ttt})$ 
\ex \label{chattermsemac} $(\lambda x.\ sleeps(x)) (\epsilon^{\Pi\alpha.\ ((\alpha\fl\ttt)\fl\alpha)} \{ani\} cat^{ani\fl \ttt})$ 
\ex \label{chattermsemred}  $sleeps^{ani\fl \ttt}  (\epsilon^{\Pi\alpha.\ ((\alpha\fl\ttt)\fl\alpha)} \{ani\} cat^{ani\fl \ttt}):\ttt$ Logical Form
\ex \label{chattermsempresup} $cat(\epsilon^{\Pi\alpha.\ ((\alpha\fl\ttt)\fl\alpha)} \{ani\} cat^{ani\fl \ttt}):\ttt$ Presupposition 
\end{xlist} 
\end{exe}

In order to apply \ma{a} to \ma{cat} a predicate of type 
$ani\fl\ttt$ 
the $\epsilon$ must be specialised to $\alpha=ani$. 
The verb \ma{sleeps} can apply to result of \ma{a cat} which is  of type $ani$, 
and the final term (\ref{chattermsemac}) is of type  $\ttt$ as expected 
  --- as explained in section 
  provided there actually exists a cat this epsilon formula with out any first order equivalent (see subsection \ref{beyond}) can be understood as   $\exists x:ani\quad sleep(x)$. 
Our analysis ought to be completed: nothing tells us that $cat(\epsilon cat)$ ($\exists x.\ cat(x)$), i.e. that a \ma{cat} actually exists ...  and this needs to be added as a presupposition.
In fact,  such a presupposition is added as soon as a determiner or an existential quantifier appears: when an utterance \ma{a cat} appears, the existence of the corresponding entity ought to be asserted. 

We use the word  \ma{presupposition} with the same sense as Asher \cite{Asher2011wow}  when he calls  \ma{presupposition} a selectional restriction:
a verb like \ma{sleeps} presupposes that its subject is an \ma{animal}. This really  is some sort of  presupposition, indeed it is quite difficult to deny a type judgement, both formally and linguistically:

\begin{itemize} 
\item Formally:  To refute ($a{:}A$) is not easy.  Indeed the complement of a type is not a type, i.e. the negation of $a{:}A$ is not $a{:}\lnot A$
--- as opposed to $\tilde A(x)$ whose negation is easily formulated as $\lnot \tilde A(x)$
\item Linguistically: If one says \ma{Rex is sleeping in the garden.}  the reply: --- \ma{No, Rex is not an animal}, that \emph{refutes a typing judgment} ($Rex{:}ani$) is difficult to utter out of the blue and needs to be better introduced and justified. On the other hand it is easy to utter an answer that \emph{refutes 
the proposition}: --- \ma{No, Rex is not sleeping, he just left.}
\end{itemize} 

\subsection{A rather satisfying account of determiners}

We started with three objections to the standard account of determiners in Montague semantics. We proposed a model that avoids those three problems:
\begin{enumerate}
\item Epsilon are individuals that can be interpreted as such (even though their interpretation does not ensure completeness of the epsilon calculus). 
\item With epsilon terms, the syntactical structure  and the structure of the logical form match. 
\item For an  indefinite determiner phrase, which  corresponds to an existential statement, there is not  anymore an irrelevant symmetry between the noun (topic, theme) and the verb phrase (comment, rheme).
\end{enumerate} 

As in von Heusinger's work, one can give a similar account of definite descriptions, the main difference being at the interpretation level: the definite description should be interpreted as the most salient entity in the context. This entity is usually introduced by an indefinite description, that is another epsilon term defined from  the same property (from the same logical formulae). The difference between a  definite description and an indefinite determiner phrase is that the former one refers to an existing discourse referent while the  later one introduces a new discourse referent. 

This also  provides a natural account of Evan's E-type pronoun \cite{Evans77pronouns}: the semantics of the pronoun \ma{he} in the example below can be copied from its antecedent to obtain the semantics of these two sentences.

\begin{exe}
\ex \label{definite} A man entered the conference hall. The man sat nearby the window. 
\ex \label{diffinterpret} A man$_1$ entered the conference hall and sat nearby the window. A man$_2$ ($\neq$ man$_1$) told him that he  just missed two slides. 
\ex \label{evans} A man entered the conference hall. He sat nearby the window. 
\end{exe} 

Universal quantification can be treated just like indefinite determiners. A universally quantified NP  corresponds to the term  $\tau_x.P(x)=\epsilon_x. \lnot(P(x))$ (c.f. section \ref{Hilbert}). The $\tau$-terms are actually  much easier to interpret than the $\epsilon$-terms: it's a generic entity with respect to property $P$. Furthermore one can introduce operators for generalised and vague quantifiers like \ma{most}, \ma{few}, \ma{a third of} etc. 

The approach to existential quantification is rather similar to choice functions that have been used in formal semantics, especially in Steedman recent book \cite{Steedman2012scope}, who also enjoy the three properties above. There are nevertheless some differences: 
\begin{itemize} 
\item The syntax, the definition of epsilon terms,  is simple. I think  different  choice functions are  needed for all the formulae, while a single epsilon is enough (and possibly already too much). 
\item Universal quantification can be treated un just the same way with $\tau_x.P(x)=\epsilon_x. \lnot(P(x))$ and even generalised and vague quantifiers can be treated that way. 
\end{itemize} 

Of course the challenging difficulty of epsilon is to find the proper notion of model which would give a completeness theorem for all the formulae including the one that do not have a first order equivalent.

\section{Conclusion} 

This work is an investigation of the outcomes of the Montagovian generative lexicon, which was designed for lexical semantics,  in formal semantics. 
The many sorted compositional framework seems to be a rich setting to explore some new direction like a typed and richer view of epsilon terms as the semantics of determiner phrases. 

We did not elaborate on scope issues: using freely the epsilon and tau operators is a form of underspecification. It involves formulae that are \emph{not} part of first order logic, like: $R(\epsilon_x P(x), \tau_z Q(z))$.  

As we showed here,  this refinement of Montague semantics draws intriguing connections between type theory --- say a judgement $a:A$ --- and many sorted logic --- a formula $\tilde A(a)$: 
we hope to understand better those issues in future work. 

As far as quantification is concerned, we would like to better understand formulae of the epsilon calculus that do not have any equivalent in usual logic and any proper notion of model, complete if possible, would help a lot.

We presently are doing psycholinguistic experiments to see how do we naturally interpret determiner phrases,  by confronting sentence to pictures in which they can be true or not, measuring reaction time and recording eye tracking. This will possibly confirm or refute the soundness of some cognitive arguments. 

The possibly to model with Hilbert operators generalised quantifiers like \ma{a third of} and vague quantifiers like \ma{many} if of course very appealing, and we already made some advances  in this direction. \cite{Retore2012rlv} Nevertheless we should not be too ambitious: basic epsilon terms already goes beyond usual first order logic, and although they do have deduction rules they lack proper models. So the situation is probably much more complicated with Hilbert terms for generalised quantifiers, which do not even have proper deductions rules. Hence such terms are a natural and appealing but mathematically difficult approach to quantification related to the semantics of determiner phrases. 

\medskip 

\noindent\textbf{Thanks} to the anonymous colleagues who provided some comments on this paper and  to Michele Abrusci, Nicholas Asher, Francis Corblin, Ulrich Kohlenbach, Zhaohui Luo, Richard Moot, Fabio Pasquali for helpful discussions. 

\bibliography{bigbiblio} 

\begin{thebibliography}{10}

\bibitem{Asher2011wow}
Asher, N.:
\newblock Lexical Meaning in context -- a web of words.
\newblock Cambridge University press (2011)

\bibitem{Asser1957}
Asser, G.:
\newblock Theorie der logischen auswahlfunktionen.
\newblock Zeitschrift f{\"u}r Mathematische Logik und Grundlagen der Mathematik
  (1957)

\bibitem{BMRjolli}
{B}assac, C., {M}ery, B., {R}etor{\'e}, C.:
\newblock {T}owards a {T}ype-{T}heoretical {A}ccount of {L}exical {S}emantics.
\newblock {J}ournal of {L}ogic {L}anguage and {I}nformation \textbf{19}(2)
  (April 2010)  229--245

\bibitem{Zbl0327.02013}
Canty, J.T.:
\newblock Zbl0327.02013 : review of ``on an extension of {H}ilbert's second
  $\epsilon$-theorem'' by {T}. {B}. {F}lanagan (jsl, 1975)

\bibitem{libera1996querelle}
de~Libera, A.:
\newblock La querelle des universaux de Platon {\`a} la fin du Moyen {\^A}ge.
\newblock Des travaux. Seuil (1996)

\bibitem{EgliHeusinger1995}
Egli, U., von Heusinger, K.:
\newblock The epsilon operator and {E}-type pronouns.
\newblock In Egli, U., Pause, P.E., Schwarze, C., von Stechow, A., Wienold, G.,
  eds.: Lexical Knowledge in the Organization of Language.
\newblock Benjamins (1995)  121--141

\bibitem{Evans77pronouns}
Evans, G.:
\newblock Pronouns, quantifiers, and relative clauses (i).
\newblock Canadian Journal of Philosophy \textbf{7}(3) (1977)  467--536

\bibitem{Girard71}
Girard, J.Y.:
\newblock Une extension de l'interpr{\'e}tation de {G}{\"o}del {\`a} l'analyse
  et son application: l'{\'e}limination des coupures dans l'analyse et la
  th{\'e}orie des types.
\newblock In Fenstad, J.E., ed.: Proceedings of the Second Scandinavian Logic
  Symposium. Volume~63 of Studies in Logic and the Foundations of Mathematics.,
  Amsterdam, North Holland (1971)  63--92

\bibitem{Girard2011blindspot}
Girard, J.Y.:
\newblock The blind spot -- lectures on logic.
\newblock European Mathematical Society (2011)

\bibitem{HBvol2}
Hilbert, D., Bernays, P.:
\newblock Grundlagen der Mathematik. Bd. 2.
\newblock Springer (1939) Traduction fran{\c c}aise de F. Gaillard, E.
  Guillaume et M. Guillaume, L'Harmattan, 2001.

\bibitem{KK86}
Kneale, W., Kneale, M.:
\newblock The development of logic. 3$^{\textrm{rd}}$ edn.
\newblock Oxford University Press (1986)

\bibitem{Leisenring1967epsilon}
Leisenring, A.C.:
\newblock Mathematical logic and Hilbert's $\epsilon$ symbol.
\newblock University Mathematical Series. Mac Donald \& Co. (1967)

\bibitem{Luo2011lacl}
Luo, Z.:
\newblock Contextual analysis of word meanings in type-theoretical semantics.
\newblock In Pogodalla, S., Prost, J.P., eds.: LACL. Volume 6736 of LNCS.,
  Springer (2011)  159--174

\bibitem{Luo2012lacl}
Luo, Z.:
\newblock Common nouns as types.
\newblock In B{\'e}chet, D., Dikovsky, A.J., eds.: LACL. Volume 7351 of Lecture
  Notes in Computer Science., Springer (2012)  173--185

\bibitem{MMR2013lenls}
Mery, B., Moot, R., Retor{\'e}, C.:
\newblock Plurals: individuals and sets in a richly typed semantics.
\newblock In Yatabe, S., ed.: Logic and Engineering of Natural Language
  Semantics 10 (LENLS 10), Keio University (2013)  143--156 ISBN
  978-4-915905-57-5.

\bibitem{MeryRetore2013nlpcs}
Mery, B., Retor\'e, C.:
\newblock Semantic types, lexical sorts and classifiers.
\newblock In Sharp, B., Zock, M., eds.: 10th International Workshop on Natural
  Language Processing and Cognitive Science, Marseilles (September 2013)

\bibitem{Zbl0381.03042}
Mints, G.:
\newblock Zbl0381.03042: review of ``cut elimination in a {G}entzen-style
  $\epsilon$-calculus without identity" by {L}inda {W}essels ({Z}. math {L}ogik
  {G}rundl. {M}ath., 1977)

\bibitem{mints2008}
Mints, G.:
\newblock Cut elimination for a simple formulation of epsilon calculus.
\newblock Ann. Pure Appl. Logic \textbf{152}(1-3) (2008)  148--160

\bibitem{moot10grail}
Moot, R.:
\newblock Wide-coverage {French} syntax and semantics using {Grail}.
\newblock In: Proceedings of Traitement Automatique des Langues Naturelles
  (TALN), Montreal (2010)

\bibitem{MootRetore2012lcg}
Moot, R., Retor{\'e}, C.:
\newblock The logic of categorial grammars: a deductive account of natural
  language syntax and semantics. Volume 6850 of LNCS.
\newblock Springer (2012)

\bibitem{MoserZach2006etheorem}
Moser, G., Zach, R.:
\newblock The epsilon calculus and herbrand complexity.
\newblock Studia Logica \textbf{82}(1) (2006)  133--155

\bibitem{RealRetore2014jolli}
Real, L., Retor{\'e}, C.:
\newblock Deverbal semantics and the {M}ontagovian generative lexicon
  ${\Lambda} \!\mathsf {{T}y}_n$.
\newblock Journal of Logic, Language and Information (2014)  1--20

\bibitem{Retore2012rlv}
Retor{\'e}, C.:
\newblock Variable types for meaning assembly: a logical syntax for generic
  noun phrases introduced by "most".
\newblock Recherches Linguistiques de Vincennes \textbf{41} (2012)  83--102

\bibitem{Retore2013taln}
Retor{\'e}, C.:
\newblock S{\'e}mantique des d{\'e}terminants dans un cadre richement typ{\'e}.
\newblock In Morin, E., Est{\`e}ve, Y., eds.: Traitement Automatique du Langage
  Naturel, TALN RECITAL 2013. Volume~1., ACL Anthology (2013)  367--380

\bibitem{Russell1905}
Russell, B.:
\newblock On denoting.
\newblock Mind \textbf{56}(14) (1905)  479--493

\bibitem{Steedman2012scope}
Steedman, M.:
\newblock Taking Scope: The Natural Semantics of Quantifiers.
\newblock MIT Press (2012)

\bibitem{Heusinger1997}
von Heusinger, K.:
\newblock Definite descriptions and choice functions.
\newblock In Akama, S., ed.: Logic, Language and Computation, Kluwer (1997)
  61--91

\bibitem{Heusinger2004}
von Heusinger, K.:
\newblock Choice functions and the anaphoric semantics of definite nps.
\newblock Research on Language and Computation \textbf{2} (2004)  309--329

\end{thebibliography}

\end{document}